\documentclass[runningheads]{llncs}
\usepackage[T1]{fontenc}
\usepackage{graphicx}
\usepackage{amsmath,amssymb}
\usepackage{booktabs}
\usepackage{multirow}
\usepackage{algorithm}
\usepackage{algorithmic}
\usepackage{hyperref}
\usepackage[table]{xcolor}
\usepackage{placeins}

\newcommand{\bard}{\textsc{bard}}

\newcommand{\vq}{\mathbf{q}}
\newcommand{\vv}{\dot{\mathbf{q}}}
\newcommand{\va}{\ddot{\mathbf{q}}}
\newcommand{\vtau}{\boldsymbol{\tau}}
\begin{document}
\title{Batched Differentiable Rigid Body Dynamics\\
in PyTorch for GPU-Accelerated Robot Learning}
\titlerunning{Batched Differentiable Rigid Body Dynamics in PyTorch}
%
% \author{Yue Wang\inst{1}\orcidID{0000-0003-2384-4639} \and
% Yanran Xu\inst{1}\orcidID{0009-0007-8892-8055} \and
% Wenbo Wu\inst{1}\orcidID{0009-0002-3937-0124} \and
% Chuanhang Qiu\inst{1}\orcidID{0000-0002-3427-4603} \and
% Zhaoxing Li \inst{1}\orcidID{0000-0003-3560-3461}}

\author{Yue Wang\inst{1}\and
Yanran Xu\inst{1}\and
Wenbo Wu\inst{1}\and
Chuanhang Qiu\inst{1} \and
Zhaoxing Li \inst{1}}
\authorrunning{Wang et al.}
\institute{University of Southampton, Southampton, SO17 1BJ, UK \\
\email{\{Yue.Wang,Y.Xu,Wenbo.Wu,C.Qiu,Zhaoxing.Li\}@soton.ac.uk}}

\maketitle
%
% ================================================================
% ABSTRACT
% ================================================================
\begin{abstract}

As robot control shifts toward large-scale reinforcement learning with in-loop dynamics computation, the community's reliance on CPU-bound libraries such as Pinocchio creates a throughput bottleneck in GPU-based training pipelines. We present \bard{} (\textbf{B}atched \textbf{A}rticulated \textbf{R}igid-body \textbf{D}ynamics), a self-contained PyTorch implementation of Featherstone's rigid-body dynamics algorithms, optimized for batched GPU evaluation and automatic differentiation. Three design choices make this efficient: a tiered lazy-evaluation cache that avoids redundant tree traversals, matmul-free joint transforms via pre-computed Rodrigues constants, and level-parallel propagation that reduces sequential operations to tree-depth batched steps. On five robot models (7--23\,DOF), \bard{} matches Pinocchio numerically while reaching up to $64\times$ higher throughput for Forward Kinematics and $63\times$ for Jacobians at batch size 4096 on an NVIDIA H200. We validate differentiability through gradient-based system identification on a 7-DOF manipulator, recovering link masses to $1.24\%$ mean error under $5\%$ torque noise, and integrate \bard{} into an Isaac Lab AMP training pipeline for an 11-DOF spined quadruped with 4096 parallel environments, where it is $8.5\times$ faster than Pinocchio and $2.0\times$ faster than ADAM for in-loop dynamics. \bard{} is open-sourced at: \textbf{\url{https://github.com/YueWang996/bard-pytorch-dynamics}}.

\end{abstract}

% ================================================================
% 1. INTRODUCTION
% ================================================================
\section{Introduction}\label{sec:intro}
With the increasing complexity of both robot structures~\cite{bledt2018cheetah,wang2025sparc,zhuang2025humanoid,8630605} and deployment environments~\cite{zhou2022swarm,sathyamoorthy2023vern,mattamala2025wild,xu2025blind,li2025hmcf}, control techniques have shifted from trajectory optimization~\cite{wang2025quattro} to large-scale reinforcement learning~\cite{rudin2021learning,hwangbo2019learning} to achieve more generalizable behavior. To reduce exploration in high-dimensional action spaces, rigid-body dynamics can be included in the training loop as a guide. For example, they can serve as differentiable models for gradient-based system identification~\cite{le2021differentiable}, or as analytical operators (Jacobians, inverse dynamics) within physics-informed reward and constraint formulations~\cite{le2021differentiable}. These applications need standalone dynamics operators (e.g.\ a Jacobian or mass-matrix call) that can be embedded in a custom loss function, not a monolithic simulator whose internals are inaccessible. Current RL training frameworks such as Isaac Lab~\cite{makoviychuk2021isaac} and MuJoCo Playground~\cite{zakka2025mujoco} use a PyTorch~\cite{paszke2019pytorch} backend with batched GPU execution for efficiency; however, the community's standard dynamics library, Pinocchio~\cite{carpentier2019pinocchio}, is CPU-bound and sequential, neither exploiting GPU parallelism nor avoiding the CPU--GPU data transfer overhead introduced at every call. Existing GPU alternatives like Brax~\cite{freeman2021brax} and MuJoCo MJX~\cite{zakka2025mujoco} provide high throughput but are tied to non-PyTorch ecosystems (JAX), making them hard to integrate into PyTorch training loops. Differentiable off-the-shelf dynamics libraries such as ADAM~\cite{lerario2025adam} sacrifice GPU throughput for backend generality (see Section~\ref{sec:related}). A native, high-performance, modular dynamics library for batched GPU evaluation within PyTorch is still missing. 

We present \bard{} (\textbf{B}atched \textbf{A}rticulated \textbf{R}igid-body \textbf{D}ynamics), a pure-PyTorch library implementing Featherstone's robot dynamics algorithms~\cite{featherstone2008rigid}, designed for batched GPU evaluation. While Featherstone's algorithms are well established, their inherently serial tree traversals resist parallelisation, making them hard to deploy in modern RL training pipelines. We address this by reorganizing traversals into level-parallel batched operations and adding a multi-tier lazy-evaluation cache that skips redundant computation across successive calls. The entire code path is compatible with \texttt{torch.compile}, allowing the PyTorch compiler to fuse the resulting operator graph. On an NVIDIA H200 at batch size 4096, \bard{} is up to 64$\times$ faster than a sequential Pinocchio baseline for forward kinematics (FK) and Jacobian evaluation. We test \bard{} on gradient-based system identification of a 7-DOF manipulator and inside an Isaac Lab RL pipeline for the SPARC quadruped~\cite{wang2025sparc}, where \bard{} achieves 8.5$\times$ higher throughput than Pinocchio and 2$\times$ higher than ADAM for in-loop dynamics computation. Our contributions are:

\begin{enumerate}
\item \bard{}, an open-source, pure-PyTorch dynamics library for batched GPU evaluation.
\item Systems-level techniques for running recursive tree computations on GPUs, including a kinematics cache, pre-allocated workspaces, and level-parallel propagation.
\item Validation on gradient-based identification and large-scale RL training with 4096 parallel environments.
\end{enumerate}

% ================================================================
% 2. RELATED WORK
% ================================================================
\section{Related Work}\label{sec:related}
We review CPU dynamics libraries, GPU simulators, and differentiable dynamics tools, focusing on why each falls short for batched PyTorch workloads.

Pinocchio~\cite{carpentier2019pinocchio} is the standard for Featherstone's spatial-vector algebra algorithms~\cite{featherstone2008rigid}, offering highly optimized single-threaded C++ routines and analytical derivatives~\cite{carpentier2018analytical}. RBDL~\cite{felis2017rbdl} and Drake~\cite{tedrake2019drake} provide similar CPU-based functionality, and downstream frameworks such as Crocoddyl~\cite{mastalli2020crocoddyl} build on these libraries for trajectory optimization. Although these libraries provide excellent performance for single-robot configurations, they suffer from two problems when embedded in GPU training loops. First, their sequential execution model requires looping over environments one by one, leaving GPU cores idle. Second, calling a CPU library from a PyTorch pipeline forces a CPU--GPU data transmission at every evaluation; this overhead grows linearly with batch size and quickly becomes a training bottleneck.

To improve throughput, frameworks such as Brax~\cite{freeman2021brax}, MuJoCo MJX~\cite{zakka2025mujoco}, and Isaac Gym~\cite{makoviychuk2021isaac} move the entire simulation onto the GPU, achieving efficient training via massive parallelism across robots. However, they are architected as monolithic end-to-end simulators that bundle contact detection, time integration, and dynamics into a single pipeline. This design makes it difficult to extract a standalone operator. For instance, when only a subsystem's dynamics is required (such as a robot arm mounted on a quadruped), an RNEA call for gravity compensation or a CRBA call for mass-matrix analysis cannot be obtained from these simulators, and therefore cannot be embedded in a custom loss function or controller to regulate that subsystem's behavior. Beyond modularity, ecosystem compatibility is a concern: MJX is built on JAX, which precludes native integration into the PyTorch training loops used by Isaac Lab and similar frameworks. DiffTaichi~\cite{hu2019difftaichi} achieves high-performance differentiable simulation but likewise requires a programming model distinct from standard PyTorch workflows.

Recent efforts have attempted to provide modular, differentiable dynamics operators within PyTorch, but they often face performance trade-offs at scale. ADAM~\cite{lerario2025adam} supports multiple backends (CPU, GPU, JAX) through a shared abstraction layer, but this generality prevents PyTorch-specific optimizations such as kernel fusion and compile-time graph capture, and its throughput falls well below hardware capacity at large batch sizes. PyTorch-Kinematics~\cite{zhong2024pytorch_kinematics} provides batched FK and Jacobians in PyTorch but omits core dynamics algorithms (RNEA, CRBA, ABA). GRiD~\cite{plancher2022grid} delivers high-performance GPU dynamics with analytical gradients, yet relies on generated CUDA kernels outside the Python autograd ecosystem, raising the integration barrier for typical PyTorch research code. Other efforts either target single-configuration workloads rather than batched training~\cite{meier2022differentiable}, or predate modern GPU-accelerated RL pipelines altogether~\cite{degrave2019differentiable,werling2021fast}. To our knowledge, existing tools have not been designed with \texttt{torch.compile} compatibility in mind, and data-dependent control flow in their implementations causes graph breaks that prevent the PyTorch compiler from fusing operations across the computation graph.

% ================================================================
% 3. IMPLEMENTATION
% ================================================================
\section{Methodology}\label{sec:impl}

\begin{figure}[t]
\centering
\includegraphics[width=0.9\textwidth]{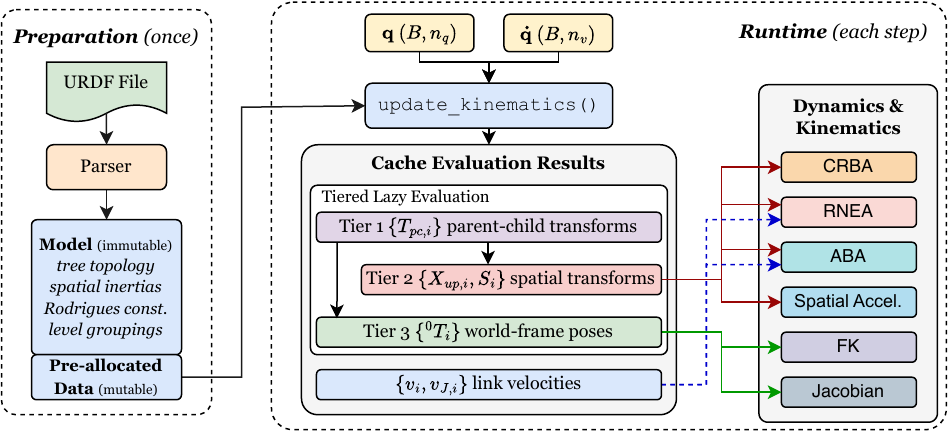}
\caption{Architecture of \bard{}. A URDF is parsed once into an immutable model storing tree topology, spatial inertias, and pre-computed constants. All mutable state for $B$ parallel environments resides in a pre-allocated workspace sized at initialisation. Downstream algorithms materialise only the kinematic quantities they require through a tiered lazy evaluation scheme.}\label{fig:architecture}
\end{figure}

\bard{} addresses these limitations by implementing the major algorithms, including forward kinematics (FK), the Recursive Newton-Euler Algorithm (RNEA), Composite Rigid Body Algorithm (CRBA), Articulated Body Algorithm (ABA), and Jacobians, as composable PyTorch operators with full \texttt{torch.compile} compatibility. Fig.~\ref{fig:architecture} shows the overall architecture. Sections~\ref{sec:lazy}--\ref{sec:jac} describe four strategies behind this performance: tiered lazy evaluation, matmul-free Rodrigues transforms, level-parallel tree propagation, and direct Jacobian column construction. Section~\ref{sec:mem} covers the underlying memory layout.

\subsection{Tiered Lazy Evaluation}\label{sec:lazy}

Different algorithms require different subsets of kinematic data. FK and Jacobian evaluation need only world-frame poses $\{{}^{0}T_i\}$, whereas RNEA, CRBA, and ABA operate on $6\times6$ spatial transforms $\{X_{\mathrm{up},i}\}$ and joint motion subspaces $\{S_i\}$. Computing all of these variables on every call would waste computation on values that are never read. \bard{} therefore splits the kinematic state into three tiers, each guarded by a validity flag and materialized only on the first downstream query that requires it:
\begin{enumerate}
    \item \textbf{Local transforms} $\{T_{\mathrm{pc},i}\}$: per-joint $4\times4$ parent-child transforms from $\vq$, shared by all downstream algorithms.
    \item \textbf{Spatial adjoints} $\{X_{\mathrm{up},i},\, X_{\mathrm{up},i}^\top,\, S_i\}$: $6\times6$ Pl\"{u}cker-frame transforms derived from the local transforms. Required only by dynamics algorithms (RNEA, CRBA, ABA).
    \item \textbf{World-frame poses} $\{{}^{0}T_i\}$: full kinematics propagation. Required only by Jacobian or FK queries over the full tree.
\end{enumerate}
As a result, an RL loop that calls only RNEA never computes world-frame propagation, and one that calls only FK never constructs spatial adjoints. Standalone FK and Jacobian functions are also provided for cases where only a single kinematic chain (e.g.\ root to end-effector) is needed, and these traverse only the relevant path without populating the full cache. 

\subsection{Matrix-Multiplication-Free Joint Transforms}\label{sec:rodrigues}

The parent-child transform for a revolute joint composes two static geometric offsets with an axis rotation:
\begin{equation}
T_{\mathrm{pc},i} = T_{\mathrm{joint},i} \cdot R_{\mathrm{axis}}(q_i) \cdot T_{\mathrm{link},i},
\end{equation}
which requires two $4\times4$ matrix multiplications per joint. However, for batched computing on a GPU, dispatching thousands of these small matmuls incurs launch overhead that far exceeds the arithmetic cost. The key observation is that the geometric offsets $T_{\mathrm{joint},i}$ and $T_{\mathrm{link},i}$ are fixed for a given robot and do not change across time or batch elements. By absorbing these constants into the Rodrigues rotation formula at model build time, we can replace the runtime matmuls with cheap element-wise operations.

Concretely, for each revolute joint with normalised axis $\hat{a}_i$ and skew matrix $K_i = [\hat{a}_i]_\times$, we pre-compute six constant matrices from the static rotations $R_{\mathrm{pre}}, t_{\mathrm{pre}}$ (of $T_{\mathrm{joint}}$) and $R_{\mathrm{post}}, t_{\mathrm{post}}$ (of $T_{\mathrm{link}}$):
\begin{align}
R^{(0)}_i &= R_{\mathrm{pre},i}\, R_{\mathrm{post},i}, &
A_i &= R_{\mathrm{pre},i}\, K_i\, R_{\mathrm{post},i}, &
B_i &= R_{\mathrm{pre},i}\, K_i^2\, R_{\mathrm{post},i}, \label{eq:rodrigues_rot}\\
t^{(0)}_i &= R_{\mathrm{pre},i}\, t_{\mathrm{post},i} + t_{\mathrm{pre},i}, &
a_i &= R_{\mathrm{pre},i}\, K_i\, t_{\mathrm{post},i}, &
b_i &= R_{\mathrm{pre},i}\, K_i^2\, t_{\mathrm{post},i}. \label{eq:rodrigues_trans}
\end{align}
At runtime, the full parent-child transform is recovered using only $\sin(q_i)$ and $\cos(q_i)$:
\begin{align}
R_{\mathrm{pc},i} &= R^{(0)}_i + \sin(q_i)\, A_i + (1-\cos (q_i))\, B_i, \label{eq:rodrigues_runtime_rot}\\
t_{\mathrm{pc},i} &= t^{(0)}_i + \sin(q_i)\, a_i + (1-\cos (q_i))\, b_i. \label{eq:rodrigues_runtime_trans}
\end{align}

The resulting expressions contain only scalar-matrix multiplies, allowing the PyTorch compiler to fuse all $n_b$ joints and $B$ environments into a single kernel launch. Prismatic and fixed joints are handled analogously with their own pre-computed constants.

\subsection{Level-Parallel Tree Propagation}\label{sec:level}

Featherstone's algorithms traverse the kinematic tree sequentially from parent to child (or child to parent), creating a data dependency chain of length $n_b$ (the number of bodies). Although a straightforward implementation processes one node at a time, sibling links at the same tree depth are in fact completely independent and can be evaluated in parallel. For example, all four hips of a quadruped can be processed simultaneously once the torso is done.

\bard{} exploits this by pre-computing a level-order grouping at build time: all nodes at the same depth are collected into a single index set, together with the memory locations of their respective parents. At runtime, each depth level is processed as one batched operation. For example, the forward velocity propagation becomes:
\begin{equation}
v_{[\ell]} = X_{\mathrm{up},[\ell]} \cdot v_{\mathrm{parent}([\ell])} + S_{[\ell]} \cdot \dot{q}_{[\ell]},
\end{equation}
where $[\ell]$ denotes all joints at depth $\ell$. This reduces $n_b$ sequential operations to $d$ (tree depth) batched ones. The same grouping is reused by RNEA's forward pass, ABA's forward acceleration pass, and the spatial-adjoint propagation in the cache. Backward passes (force and inertia accumulation from leaves to root) remain sequential in reverse topological order, since multiple children may accumulate into the same parent and cannot be safely parallelized without serializing memory access. Nevertheless, all $B$ environments are still processed simultaneously at every node, so the sequential traversal adds only $n_b$ serial steps regardless of batch size.

\subsection{Direct Jacobian Column Construction}\label{sec:jac}

The standard approach to geometric Jacobian computation forms the $6\times6$ adjoint $\mathrm{Ad}({}^{0}T_k)$ and multiplies by the joint motion subspace $S_k$ for each joint along the chain. For single-DOF joints---the vast majority in practice---this produces a single $6\times1$ column, making the full adjoint construction wasteful. For a revolute joint with axis $\hat{a}_k$, the Jacobian column can be written directly as:
\begin{equation}\label{eq:jac_direct}
J_k = \begin{bmatrix} p_k^w \times (R_k^w \, \hat{a}'_k) \\ R_k^w \, \hat{a}'_k \end{bmatrix},
\end{equation}
where $\hat{a}'_k = R_{\mathrm{offset},k}\, \hat{a}_k$ is pre-computed at build time, and $R_k^w$ and $p_k^w$ are the world-frame rotation and position already available in the FK cache. This replaces $n$ adjoint constructions with cross products and matrix-vector multiplies.

\subsection{Memory Layout and Compiler Compatibility}\label{sec:mem}

All workspace tensors use a batch-first $(B, n_b, \ldots)$ layout for coalesced GPU memory access. The transpose $X_{\mathrm{up},i}^\top$, which appears in every dynamics backward pass, is stored contiguously alongside $X_{\mathrm{up},i}$ rather than computed as a non-contiguous view, avoiding a common source of cache-line waste. For the bottleneck $6\times6$ update $\mathrm{Out} \mathrel{+}= X_i^\top M_i X_i$ that dominates CRBA and ABA, an optional fused Triton~\cite{tillet2019triton} kernel replaces two separate \texttt{bmm} calls with a single kernel launch, keeping all intermediates in registers.

\bard{} is fully compatible with \texttt{torch.compile}. To achieve this, tree metadata is stored as constant index tensors rather than Python data structures, all control flow is static (no data-dependent branching, no \texttt{.item()} calls), and critical helpers are pre-compiled. These constraints ensure that the PyTorch compiler can trace the entire forward pass as a single fused graph, enabling cross-operator optimization that would be blocked by graph breaks. For RL rollouts where gradients are not needed, \bard{} uses an in-place code path that skips gradient tracking for maximum throughput. When gradients are required (e.g.\ for system identification), the same algorithms switch to a functional mode that supports full automatic differentiation through all computations.

% ================================================================
% 4. EXPERIMENTS
% ================================================================
\section{Experiments}\label{sec:experiments}

We evaluate \bard{} in four stages: numerical agreement with Pinocchio (Section~\ref{sec:accuracy}), computational throughput across batch sizes and hardware (Section \ref{sec:speed}), gradient-based system identification (Section \ref{sec:sysid}), and integration into a large-scale RL training pipeline (Section~\ref{sec:rl}).

\subsection{Numerical Agreement}\label{sec:accuracy}

\begin{table}[t]
\caption{Maximum absolute discrepancy between \bard{} and Pinocchio in \texttt{float64}. Both libraries evaluate 1000 random configurations per model.}\label{tab:accuracy}
\centering
\setlength{\tabcolsep}{4pt}
\begin{tabular}{l@{\hspace{8pt}}c@{\hspace{8pt}}c@{\hspace{8pt}}c@{\hspace{8pt}}c@{\hspace{8pt}}c}
\toprule
Algorithm & xArm7 (7) & SPARC (11) & Go2 (12) & H1 (19) & G1 (23) \\
\midrule
FK       & $7.8\mathrm{e}{-16}$ & $8.9\mathrm{e}{-16}$ & $7.8\mathrm{e}{-16}$ & $1.0\mathrm{e}{-15}$ & $1.3\mathrm{e}{-15}$ \\
Jacobian & $2.4\mathrm{e}{-15}$ & $2.1\mathrm{e}{-15}$ & $1.8\mathrm{e}{-15}$ & $3.1\mathrm{e}{-15}$ & $2.2\mathrm{e}{-15}$ \\
RNEA     & $1.6\mathrm{e}{-14}$ & $5.0\mathrm{e}{-14}$ & $1.4\mathrm{e}{-13}$ & $5.1\mathrm{e}{-13}$ & $4.6\mathrm{e}{-13}$ \\
CRBA     & $2.0\mathrm{e}{-15}$ & $1.8\mathrm{e}{-15}$ & $3.6\mathrm{e}{-15}$ & $1.4\mathrm{e}{-14}$ & $1.4\mathrm{e}{-14}$ \\
ABA      & $2.9\mathrm{e}{-11}$ & $1.2\mathrm{e}{-10}$ & $2.3\mathrm{e}{-12}$ & $6.4\mathrm{e}{-12}$ & $5.8\mathrm{e}{-11}$ \\
\bottomrule
\end{tabular}
\end{table}

To verify correctness, we compare \bard{} against Pinocchio across all five algorithms in \texttt{float64}, using five robot models of varying complexity: xArm7 (7-DOF), SPARC~\cite{wang2025sparc} (11-DOF), Go2 (12-DOF), H1 (19-DOF), and G1 (23-DOF). For each model and algorithm, we generate 1000 random configurations and report the maximum absolute discrepancy.

Table~\ref{tab:accuracy} shows the results. FK and Jacobian discrepancies are near machine epsilon ($\approx 10^{-16}$), and RNEA and CRBA remain below $10^{-13}$, confirming that the Rodrigues-based transform computation and level-parallel propagation introduce no numerical deviation beyond floating-point rounding. ABA shows slightly larger discrepancies (up to $10^{-10}$) due to its three-pass structure accumulating more rounding error, but still within acceptable bounds for all practical applications.

\subsection{Computational Throughput}\label{sec:speed}

All throughput experiments use PyTorch 2.8 (CUDA 12.9) on an NVIDIA H200 GPU, with Pinocchio 3.7 as the CPU baseline running single-threaded in a sequential loop, which is how it is typically called within PyTorch pipelines. We benchmark \bard{} against four baselines across batch sizes from 64 to 65{,}536: (1)~Pinocchio C++ with NumPy I/O (raw C++ performance as a reference), (2)~Pinocchio with PyTorch tensor conversion (realistic pipeline cost), (3)~ADAM \cite{lerario2025adam}, the closest PyTorch-based dynamics library, and (4) the proposed~\bard{}.

\begin{figure}[t]
\centering
\includegraphics[width=\textwidth]{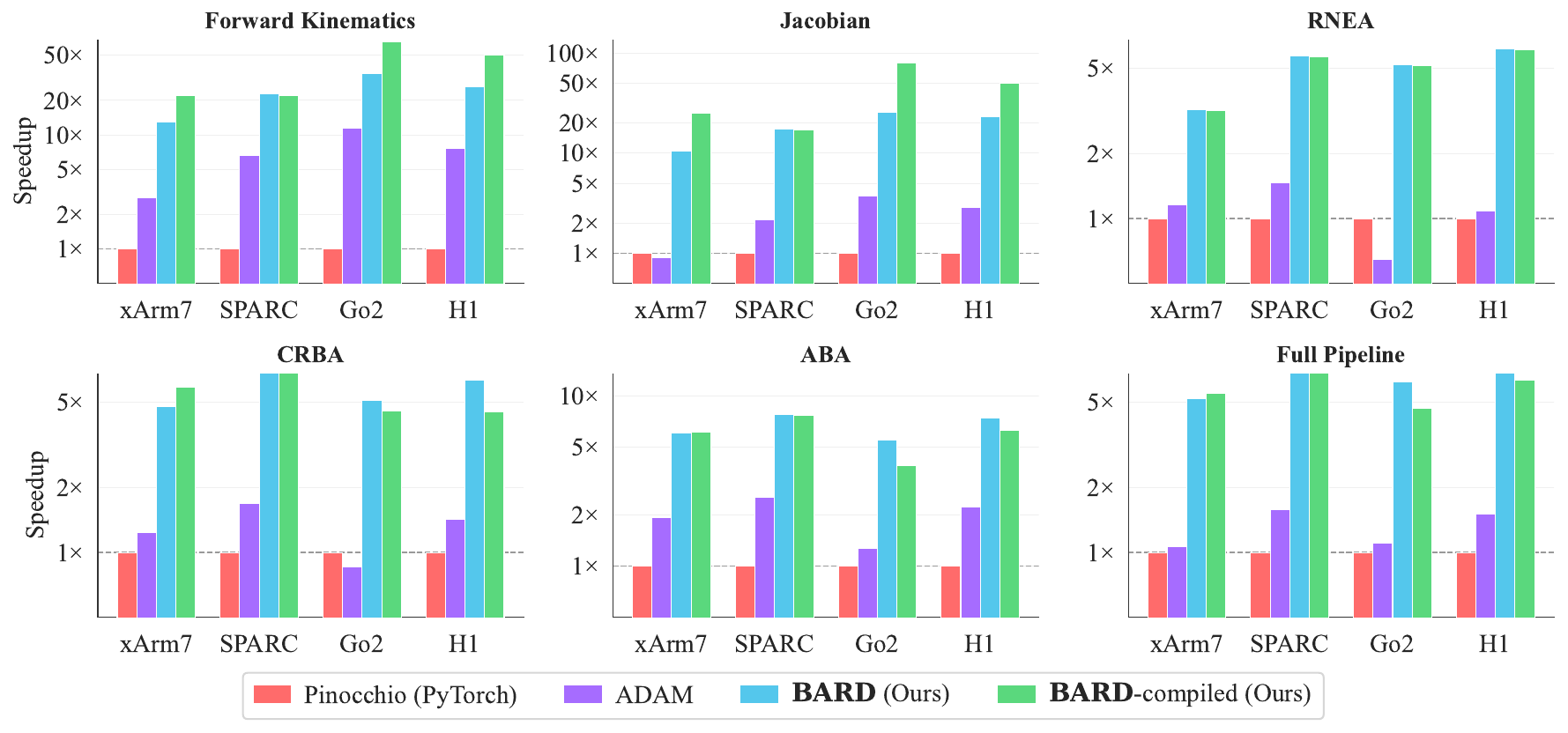}
\caption{Speedup over Pinocchio (PyTorch) at batch size 4096 on NVIDIA H200. \bard{} consistently outperforms ADAM across all algorithms, with the largest gains in FK ($64\times$) and Jacobian ($63\times$).}\label{fig:bar}
\end{figure}

Fig.~\ref{fig:bar} shows speedups at batch size 4096 on the H200. \bard{} outperforms both Pinocchio and ADAM across all algorithms, with the largest gains in kinematics: up to $64\times$ for FK and $63\times$ for Jacobian over the Pinocchio PyTorch wrapper. For dynamics algorithms the speedup is smaller but still consistent, ranging from $3\times$ to $7\times$ depending on the robot and algorithm.

\begin{figure}[t]
\centering
\includegraphics[width=\textwidth]{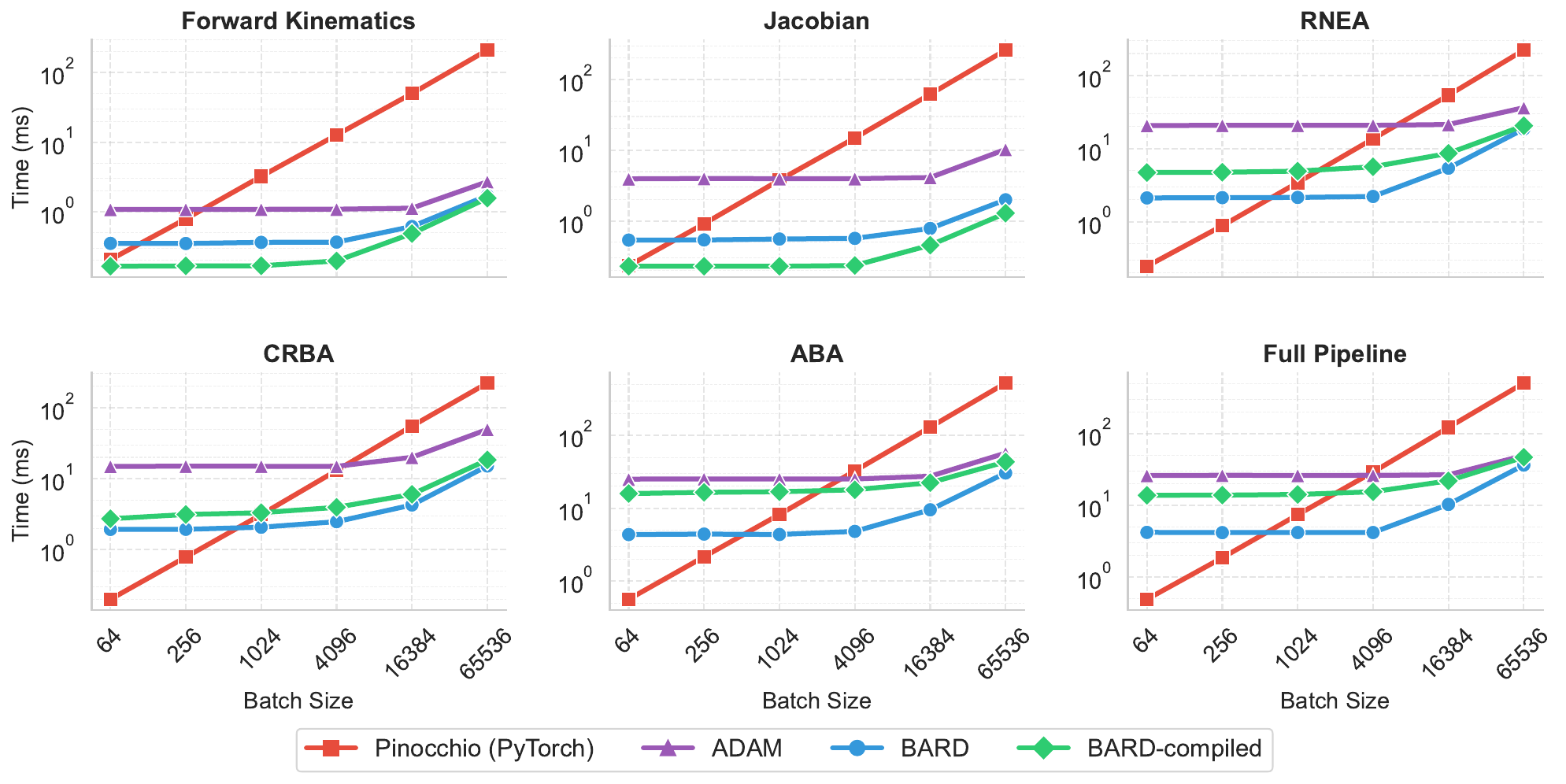}
\caption{Absolute wall-clock time vs.\ batch size for the Go2 (12-DOF) on NVIDIA H200. Pinocchio's execution time scales linearly with batch size, while \bard{} remains nearly flat up to batch size 4096.}\label{fig:scaling}
\end{figure}

Fig.~\ref{fig:scaling} shows scaling behavior on the Go2 robot. Pinocchio's wall-clock time grows linearly with batch size due to sequential CPU evaluation. \bard{}'s execution time remains nearly constant from batch size 64 to 4096 for kinematics, and rises gradually for dynamics algorithms as the computation per call is heavier. Beyond batch size 4096, all \bard{} timings begin to increase but remain over an order of magnitude faster than Pinocchio. ADAM's execution time is stable across batch sizes, but at a consistently higher level than \bard{} due to the lack of caching and kernel-level optimizations.

Notably, \texttt{torch.compile} has different effects depending on the algorithm class. For kinematics (FK, Jacobian), compilation consistently improves throughput across all GPUs and robot models we tested, as the compiler fuses the chain of element-wise Rodrigues operations into fewer kernel launches. For dynamics (RNEA, CRBA, ABA), the picture is different. On the H200, compilation produces marginal speedup for low-DOF robots (e.g.\ $3.5\times \to 3.4\times$ for RNEA on the 7-DOF xArm7) and no meaningful change for higher-DOF models. On the L4, compilation actively degrades dynamics performance: for example, RNEA on the Go2 drops from $3.0\times$ to $1.1\times$ speedup over Pinocchio, and the full pipeline drops from $4.3\times$ to $0.6\times$, making compiled \bard{} slower than the Pinocchio baseline. This occurs because the Triton backend attempts to fuse across the sequential tree traversals in dynamics algorithms, producing kernels that exceed the L4's register file and spill to slower memory. The L4 has significantly fewer streaming multiprocessors than the H200 (58 vs.\ 132) and an order of magnitude less memory bandwidth (300\,GB/s vs.\ 4.8\,TB/s), making it particularly sensitive to such register pressure. In contrast, \bard{}'s eager-mode dynamics already use hand-written Triton kernels sized for the $6\times6$ updates, avoiding this problem. In practice, we recommend eager mode for dynamics-heavy workloads on lower-bandwidth GPUs and compiled mode for kinematics-heavy workloads.

\subsection{System Identification}\label{sec:sysid}

\begin{table}[t]
\caption{System identification results on the KUKA iiwa (7-DOF). Link masses perturbed by $\pm 30\%$ are recovered via gradient-based optimization through \bard{}'s differentiable RNEA. Results averaged over 10 random seeds.}\label{tab:sysid}
\centering
\begin{tabular}{lccc}
\toprule
Link & True Mass (kg) & Identified Mass (kg) & Error (\%) \\
\midrule
link\_2 & 4.000 & $3.879 \pm 0.224$ & $4.99 \pm 3.73$ \\
link\_3 & 3.000 & $3.014 \pm 0.018$ & $0.55 \pm 0.54$ \\
link\_4 & 2.700 & $2.708 \pm 0.019$ & $0.57 \pm 0.49$ \\
link\_5 & 1.700 & $1.697 \pm 0.013$ & $0.50 \pm 0.54$ \\
link\_6 & 1.800 & $1.800 \pm 0.010$ & $0.41 \pm 0.37$ \\
link\_7 & 0.300 & $0.300 \pm 0.002$ & $0.42 \pm 0.27$ \\
\midrule
\multicolumn{3}{l}{Overall mean error} & $1.24 \pm 0.78$ \\
\bottomrule
\end{tabular}
\end{table}

To validate \bard{}'s differentiability, we apply it to gradient-based system identification, recovering link masses of the KUKA LBR iiwa (7-DOF) by minimizing
\[
\min_\theta \sum\nolimits_{k} \| \text{RNEA}(\vq^{(k)}, \vv^{(k)}, \va^{(k)}; \theta) - \vtau^{(k)}_{\text{meas}} \|^2.
\]
Ground-truth torques are generated by Pinocchio from 10{,}000 random configurations with 5\% Gaussian noise. All link masses are perturbed by $\pm 30\%$ and registered as learnable parameters, with spatial inertias rebuilt at each iteration so that autograd can differentiate through the entire pipeline from mass parameters to predicted torques without requiring explicit derivation of gradients. optimization uses L-BFGS over 10 random seeds.

Table~\ref{tab:sysid} reports the results. Convergence takes 5--10 L-BFGS iterations, yielding $1.24\% \pm 0.78\%$ mean mass error across all links and seeds. The proximal link~2 shows the highest variance, consistent with classical identifiability analysis where base-proximal parameters are harder to isolate.

\subsection{RL Training with In-Loop Dynamics}\label{sec:rl}
To demonstrate practical integration, we embed \bard{} into an Adversarial Motion Priors (AMP)~\cite{peng2021amp} training pipeline in Isaac Lab 5.1~\cite{makoviychuk2021isaac} for the SPARC quadruped. The robot features a 3-DOF actuated spine driven by a model-based task-space impedance controller. The controller calls FK, Jacobian, and RNEA at every environment step to compute spine torques, while the RL policy learns leg joint actions and spine impedance parameters. We train with 4096 parallel environments on a single NVIDIA L40S GPU and compare three dynamics backends: Pinocchio with a PyTorch wrapper, ADAM, and \bard{}. \bard{} achieves 16.75\,it/s, a $2.0\times$ speedup over ADAM (8.2\,it/s) and $8.5\times$ over the Pinocchio with PyTorch wrapper (1.96\,it/s). The three backends are numerically equivalent for the computed torques, so the choice of dynamics library affects only training wall-clock time, confirming that \bard{} is a drop-in replacement requiring no algorithmic changes.

\section{Conclusion}\label{sec:conclusion}
We presented \bard{}, a pure-PyTorch library for batched, differentiable rigid-body dynamics. By restructuring Featherstone's algorithms for GPU execution through lazy evaluation, level-parallel tree traversal, and matmul-free joint transforms, \bard{} achieves numerical accuracy matching Pinocchio while delivering up to $64\times$ throughput improvement at batch size 4096. Gradient-based system identification recovers link masses to 1.24\% error, and integration into an Isaac Lab AMP pipeline for the SPARC robot achieves $2.0\times$ higher training throughput than ADAM and $8.5\times$ over Pinocchio, making large-scale RL training with in-loop dynamics practical on a single GPU. The throughput advantage is largest for kinematics-dominated workloads at large batch sizes; for small kinematic chains (e.g.\ 3-DOF), per-kernel launch overhead narrows the gap, as reflected by the more modest $2\times$ speedup over ADAM in the RL experiment compared with the $64\times$ seen in pure-kinematics benchmarks. Additionally, as shown in Section~\ref{sec:speed}, \texttt{torch.compile} can degrade dynamics performance on smaller GPUs due to register spilling, so users should profile before enabling compilation for dynamics-heavy workloads. Future work includes contact dynamics and PyTorch-native trajectory optimization, where \bard{}'s differentiable RNEA could serve as the dynamics backend for trajectory optimization solvers.

% ================================================================
% CREDITS
% ================================================================
% \begin{credits}
% \subsubsection{\discintname}
% The author has no competing interests to declare that are relevant to the content of this article.
% \end{credits}

% ================================================================
% BIBLIOGRAPHY
% ================================================================
\bibliographystyle{splncs04}
\bibliography{references}

% ================================================================
% SUPPLEMENTARY MATERIAL (Appendix)
% ================================================================
% ================================================================
% APPENDIX: Supplementary Material
% ================================================================
\newpage
\appendix

\section*{Supplementary Material}
\addcontentsline{toc}{section}{Supplementary Material}

This appendix provides extended benchmark results for \bard{} across four NVIDIA GPUs: H100 80GB HBM3, A100 80GB PCIe, L40S, and L4.  All timings use the same protocol as the main paper: median of 100 runs at batch size 4096, with five robot models of increasing DOF.  Methods compared: Pinocchio C++ (with NumPy I/O), Pinocchio with PyTorch tensor conversion (baseline for speedup), ADAM~\cite{lerario2025adam}, \bard{} (eager), and \bard{} with \texttt{torch.compile} (\bard{}-compiled).  ADAM results for the G1 robot are unavailable due to compatibility issues with ADAM's URDF parser.

% ================================================================
% A. CROSS-GPU SPEEDUP TABLES
% ================================================================
\section{Cross-GPU Throughput Benchmarks}\label{sec:tables}

Tables~\ref{tab:speed_h100}--\ref{tab:speed_l4} report speedup over the Pinocchio (PyTorch) baseline at batch size 4096 across all five robots and six algorithm categories.  Each value is the median of 100 paired-sample ratios; brackets show the interquartile range [Q1--Q3] where it exceeds 5\% of the median (most entries have negligible variance).  Shaded cells indicate cases where \texttt{torch.compile} produces slower execution than eager mode (see Section~\ref{sec:compile} for analysis).  Bold entries denote speedups $\geq 10\times$.

% --- H100 ---
\begin{table}[htbp]
\caption{Speedup over Pinocchio (PyTorch) at batch size 4096 on NVIDIA H100 80GB HBM3. Brackets show [Q1--Q3] when IQR exceeds 5\% of the median. Shaded cells indicate \texttt{torch.compile} regression.}\label{tab:speed_h100}
\centering
\setlength{\tabcolsep}{3pt}
\small
\begin{tabular}{llrrrrrr}
\toprule
Robot & Method & FK & Jac. & RNEA & CRBA & ABA & Full \\
\midrule
\multirow{4}{*}{KUKA iiwa (7)} & Pin. C++ & 4.0$\times$ & 1.7$\times$ & 1.7$\times$ & 2.1$\times$ & 1.2$\times$ & 2.1$\times$ \\
 & ADAM & 2.6$\times$ & 0.8$\times$ & 1.0$\times$ & 1.1$\times$ & 2.0$\times$ & 1.0$\times$ \\
 & BARD & \textbf{12}$\times$ & 8.8$\times$ & 3.5$\times$ & 4.2$\times$ & 5.7$\times$ & 5.4$\times$ \\
 & BARD-c & \textbf{23}$\times$ & \textbf{23}$\times$ & 3.4$\times$ & 4.1$\times$ & 5.6$\times$ & 6.0$\times$ \\
\midrule
\multirow{4}{*}{SPARC (11)} & Pin. C++ & 4.2$\times$ & 2.0$\times$ & 2.0$\times$ & 2.6$\times$ & 1.3$\times$ & 2.1$\times$ \\
 & ADAM & 5.7$\times$ & 1.8$\times$ & 1.2$\times$ & 1.4$\times$ & 2.3$\times$ & 1.4$\times$ \\
 & BARD & \textbf{21}$\times$ & \textbf{15}$\times$ & 5.6$\times$ & 6.4$\times$ & 7.7$\times$ & 7.0$\times$ \\
 & BARD-c & \cellcolor{gray!20}\textbf{17}$\times$ & \cellcolor{gray!20}\textbf{15}$\times$ & 5.4$\times$ & 6.1$\times$ & 7.6$\times$ & 6.7$\times$ \\
\midrule
\multirow{4}{*}{Go2 (12)} & Pin. C++ & 3.0$\times$ & 1.8$\times$ & 1.9$\times$ & 2.8$\times$ & 1.3$\times$ & 1.9$\times$ \\
 & ADAM & \textbf{11}$\times$ & 3.4$\times$ & 0.5$\times$ & 0.8$\times$ & 1.2$\times$ & 1.0$\times$ \\
 & BARD & \textbf{34}$\times$ & \textbf{25}$\times$ & 5.7$\times$ & 6.5$\times$ & 7.1$\times$ & 6.7$\times$ \\
 & BARD-c & \textbf{68}$\times$ & \textbf{73}$\times$ & \cellcolor{gray!20}5.2$\times$ & 6.3$\times$ & 6.8$\times$ & \cellcolor{gray!20}5.2$\times$ \\
\midrule
\multirow{4}{*}{H1 (19)} & Pin. C++ & 2.8$\times$ & 1.7$\times$ & 1.6$\times$ & 2.2$\times$ & 1.3$\times$ & 1.7$\times$ \\
 & ADAM & 6.8$\times$ & 2.5$\times$ & 0.9$\times$ & 1.2$\times$ & 2.0$\times$ & 1.3$\times$ \\
 & BARD & \textbf{25}$\times$ & \textbf{22}$\times$ & 6.2$\times$ & 5.1$\times$ & 8.3$\times$ & 7.5$\times$ \\
 & BARD-c & \textbf{52}$\times$ & \textbf{55}$\times$ & 6.0$\times$ & 5.0$\times$ & 8.2$\times$ & \cellcolor{gray!20}6.9$\times$ \\
\midrule
\multirow{4}{*}{G1 (23)} & Pin. C++ & 2.5$\times$ & 1.6$\times$ & 1.6$\times$ & 2.1${\scriptstyle [2.0\text{--}2.2]}\times$ & 1.2$\times$ & 1.6$\times$ \\
 & ADAM & --- & --- & --- & --- & --- & --- \\
 & BARD & \textbf{24}$\times$ & \textbf{21}$\times$ & 6.1$\times$ & 4.7${\scriptstyle [4.4\text{--}4.8]}\times$ & 8.2$\times$ & 7.0$\times$ \\
 & BARD-c & \textbf{50}$\times$ & \textbf{52}$\times$ & 6.0$\times$ & 4.6${\scriptstyle [4.4\text{--}4.7]}\times$ & 8.1$\times$ & \cellcolor{gray!20}5.6$\times$ \\
\bottomrule
\end{tabular}
\end{table}

% --- A100 ---
\begin{table}[htbp]
\caption{Speedup over Pinocchio (PyTorch) at batch size 4096 on NVIDIA A100 80GB PCIe. Brackets show [Q1--Q3] when IQR exceeds 5\% of the median. Shaded cells indicate \texttt{torch.compile} regression.}\label{tab:speed_a100}
\centering
\setlength{\tabcolsep}{3pt}
\small
\begin{tabular}{llrrrrrr}
\toprule
Robot & Method & FK & Jac. & RNEA & CRBA & ABA & Full \\
\midrule
\multirow{4}{*}{KUKA iiwa (7)} & Pin. C++ & 4.7$\times$ & 1.7$\times$ & 1.7$\times$ & 2.2$\times$ & 1.3$\times$ & 2.2$\times$ \\
 & ADAM & 2.3$\times$ & 0.7$\times$ & 0.9$\times$ & 1.0$\times$ & 1.5$\times$ & 0.8$\times$ \\
 & BARD & 9.8$\times$ & 7.2$\times$ & 2.9$\times$ & 3.4$\times$ & 4.3$\times$ & 4.6$\times$ \\
 & BARD-c & \textbf{21}$\times$ & \textbf{21}$\times$ & 2.9$\times$ & 3.3$\times$ & 4.3$\times$ & 5.5$\times$ \\
\midrule
\multirow{4}{*}{SPARC (11)} & Pin. C++ & 5.0$\times$ & 2.0$\times$ & 2.0$\times$ & 2.6$\times$ & 1.4$\times$ & 2.1$\times$ \\
 & ADAM & 4.9$\times$ & 1.5$\times$ & 1.0$\times$ & 1.3$\times$ & 1.8$\times$ & 1.1$\times$ \\
 & BARD & \textbf{17}$\times$ & \textbf{13}$\times$ & 4.7$\times$ & 5.1$\times$ & 5.9$\times$ & 5.8$\times$ \\
 & BARD-c & \cellcolor{gray!20}\textbf{16}$\times$ & \textbf{13}$\times$ & 4.7$\times$ & 5.0$\times$ & 5.9$\times$ & 5.7$\times$ \\
\midrule
\multirow{4}{*}{Go2 (12)} & Pin. C++ & 3.8$\times$ & 1.9$\times$ & 1.9$\times$ & 2.7$\times$ & 1.3$\times$ & 2.0$\times$ \\
 & ADAM & 8.6$\times$ & 2.7$\times$ & 0.4$\times$ & 0.7$\times$ & 0.9$\times$ & 0.8$\times$ \\
 & BARD & \textbf{26}$\times$ & \textbf{20}$\times$ & 4.8$\times$ & 4.9$\times$ & 5.2$\times$ & 5.2$\times$ \\
 & BARD-c & \textbf{55}$\times$ & \textbf{59}$\times$ & \cellcolor{gray!20}4.3$\times$ & 4.9$\times$ & 4.9$\times$ & \cellcolor{gray!20}4.1$\times$ \\
\midrule
\multirow{4}{*}{H1 (19)} & Pin. C++ & 3.3$\times$ & 1.7$\times$ & 1.7$\times$ & 2.5${\scriptstyle [2.3\text{--}2.5]}\times$ & 1.3$\times$ & 1.8$\times$ \\
 & ADAM & 5.5$\times$ & 2.0$\times$ & 0.7$\times$ & 1.1$\times$ & 1.5$\times$ & 1.1$\times$ \\
 & BARD & \textbf{20}$\times$ & \textbf{17}$\times$ & 5.2$\times$ & 4.3$\times$ & 6.0$\times$ & 6.1$\times$ \\
 & BARD-c & \textbf{44}$\times$ & \textbf{45}$\times$ & 5.1$\times$ & 4.3$\times$ & 5.9$\times$ & \cellcolor{gray!20}5.7$\times$ \\
\midrule
\multirow{4}{*}{G1 (23)} & Pin. C++ & 2.9$\times$ & 1.7$\times$ & 1.6$\times$ & 2.4${\scriptstyle [2.2\text{--}2.4]}\times$ & 1.2$\times$ & 1.7$\times$ \\
 & ADAM & --- & --- & --- & --- & --- & --- \\
 & BARD & \textbf{19}$\times$ & \textbf{17}$\times$ & 5.0$\times$ & 3.9${\scriptstyle [3.6\text{--}4.0]}\times$ & 5.8$\times$ & 5.6$\times$ \\
 & BARD-c & \textbf{43}$\times$ & \textbf{43}$\times$ & 4.9$\times$ & 3.8${\scriptstyle [3.6\text{--}3.9]}\times$ & 5.8$\times$ & \cellcolor{gray!20}4.6$\times$ \\
\bottomrule
\end{tabular}
\end{table}

% --- L40S ---
\begin{table}[htbp]
\caption{Speedup over Pinocchio (PyTorch) at batch size 4096 on NVIDIA L40S. Brackets show [Q1--Q3] when IQR exceeds 5\% of the median. Shaded cells indicate \texttt{torch.compile} regression.}\label{tab:speed_l40s}
\centering
\setlength{\tabcolsep}{3pt}
\small
\begin{tabular}{llrrrrrr}
\toprule
Robot & Method & FK & Jac. & RNEA & CRBA & ABA & Full \\
\midrule
\multirow{4}{*}{KUKA iiwa (7)} & Pin. C++ & 4.3$\times$ & 1.6$\times$ & 1.6$\times$ & 2.1$\times$ & 1.3$\times$ & 2.0$\times$ \\
 & ADAM & 3.4$\times$ & 1.1$\times$ & 1.4$\times$ & 1.4$\times$ & 2.4$\times$ & 1.3$\times$ \\
 & BARD & \textbf{14}$\times$ & \textbf{12}$\times$ & 4.5$\times$ & 5.4$\times$ & 7.2$\times$ & 6.7$\times$ \\
 & BARD-c & \textbf{26}$\times$ & \textbf{31}$\times$ & 4.5$\times$ & 5.2$\times$ & 7.1$\times$ & 7.8$\times$ \\
\midrule
\multirow{4}{*}{SPARC (11)} & Pin. C++ & 4.5$\times$ & 1.9$\times$ & 1.9$\times$ & 2.3$\times$ & 1.3$\times$ & 2.0$\times$ \\
 & ADAM & 7.3$\times$ & 2.4$\times$ & 1.6$\times$ & 1.8$\times$ & 2.8$\times$ & 1.8$\times$ \\
 & BARD & \textbf{25}$\times$ & \textbf{19}$\times$ & 7.1$\times$ & 7.9$\times$ & 9.2$\times$ & 8.5$\times$ \\
 & BARD-c & \cellcolor{gray!20}\textbf{22}$\times$ & \textbf{19}$\times$ & 7.0$\times$ & 7.6$\times$ & 9.1$\times$ & 8.3$\times$ \\
\midrule
\multirow{4}{*}{Go2 (12)} & Pin. C++ & 3.4$\times$ & 1.8$\times$ & 1.8$\times$ & 2.3$\times$ & 1.3$\times$ & 1.8$\times$ \\
 & ADAM & \textbf{13}$\times$ & 4.2$\times$ & 0.7$\times$ & 0.9$\times$ & 1.3$\times$ & 1.2$\times$ \\
 & BARD & \textbf{38}$\times$ & \textbf{30}$\times$ & 6.9$\times$ & 5.5$\times$ & 8.1$\times$ & 7.8$\times$ \\
 & BARD-c & \textbf{73}$\times$ & \textbf{90}$\times$ & \cellcolor{gray!20}3.2$\times$ & 5.5$\times$ & \cellcolor{gray!20}5.1$\times$ & \cellcolor{gray!20}2.2$\times$ \\
\midrule
\multirow{4}{*}{H1 (19)} & Pin. C++ & 3.0$\times$ & 1.6$\times$ & 1.6$\times$ & 2.2$\times$ & 1.2$\times$ & 1.7$\times$ \\
 & ADAM & 8.6$\times$ & 3.2$\times$ & 1.2$\times$ & 1.6$\times$ & 2.4$\times$ & 1.7$\times$ \\
 & BARD & \textbf{30}$\times$ & \textbf{26}$\times$ & 8.3$\times$ & 6.8$\times$ & \textbf{10}$\times$ & 9.3$\times$ \\
 & BARD-c & \textbf{59}$\times$ & \textbf{65}$\times$ & \cellcolor{gray!20}4.5$\times$ & 6.7$\times$ & \cellcolor{gray!20}7.3$\times$ & \cellcolor{gray!20}4.1$\times$ \\
\midrule
\multirow{4}{*}{G1 (23)} & Pin. C++ & 2.6$\times$ & 1.6$\times$ & 1.5$\times$ & 2.0$\times$ & 1.2$\times$ & 1.5$\times$ \\
 & ADAM & --- & --- & --- & --- & --- & --- \\
 & BARD & \textbf{28}$\times$ & \textbf{25}$\times$ & 7.4$\times$ & 5.6$\times$ & 9.7$\times$ & 8.5$\times$ \\
 & BARD-c & \textbf{54}$\times$ & \textbf{60}$\times$ & 7.4$\times$ & 5.5$\times$ & 9.6$\times$ & \cellcolor{gray!20}2.8$\times$ \\
\bottomrule
\end{tabular}
\end{table}

% --- L4 ---
\begin{table}[htbp]
\caption{Speedup over Pinocchio (PyTorch) at batch size 4096 on NVIDIA L4. Brackets show [Q1--Q3] when IQR exceeds 5\% of the median. Shaded cells indicate \texttt{torch.compile} regression.}\label{tab:speed_l4}
\centering
\setlength{\tabcolsep}{3pt}
\small
\begin{tabular}{llrrrrrr}
\toprule
Robot & Method & FK & Jac. & RNEA & CRBA & ABA & Full \\
\midrule
\multirow{4}{*}{KUKA iiwa (7)} & Pin. C++ & 4.1$\times$ & 1.6$\times$ & 1.6$\times$ & 2.0$\times$ & 1.2$\times$ & 2.0$\times$ \\
 & ADAM & 3.3$\times$ & 1.1$\times$ & 1.4$\times$ & 0.9$\times$ & 2.0$\times$ & 1.2$\times$ \\
 & BARD & \textbf{12}$\times$ & 8.8$\times$ & 4.1$\times$ & 4.6$\times$ & 6.8$\times$ & 6.4$\times$ \\
 & BARD-c & \textbf{14}$\times$ & \textbf{15}$\times$ & 4.1$\times$ & 4.6$\times$ & 6.7$\times$ & \cellcolor{gray!20}5.5$\times$ \\
\midrule
\multirow{4}{*}{SPARC (11)} & Pin. C++ & 4.4$\times$ & 1.9$\times$ & 1.9$\times$ & 2.3$\times$ & 1.3$\times$ & 1.9$\times$ \\
 & ADAM & 7.2$\times$ & 2.4$\times$ & 1.6$\times$ & 1.3$\times$ & 2.4$\times$ & 1.7$\times$ \\
 & BARD & \textbf{23}$\times$ & \textbf{16}$\times$ & 6.1$\times$ & 5.9$\times$ & 8.3$\times$ & 7.9$\times$ \\
 & BARD-c & \cellcolor{gray!20}\textbf{21}$\times$ & \textbf{16}$\times$ & 6.1$\times$ & 5.8$\times$ & 8.2$\times$ & 7.8$\times$ \\
\midrule
\multirow{4}{*}{Go2 (12)} & Pin. C++ & 3.4$\times$ & 1.8$\times$ & 1.8$\times$ & 2.3$\times$ & 1.3$\times$ & 1.8$\times$ \\
 & ADAM & \textbf{13}$\times$ & 4.2$\times$ & 0.7$\times$ & 0.6$\times$ & 1.1$\times$ & 1.2$\times$ \\
 & BARD & \textbf{35}$\times$ & \textbf{25}$\times$ & 3.0$\times$ & 2.8$\times$ & 4.2$\times$ & 4.3$\times$ \\
 & BARD-c & \textbf{43}$\times$ & \textbf{37}$\times$ & \cellcolor{gray!20}1.1$\times$ & 2.8$\times$ & \cellcolor{gray!20}2.1$\times$ & \cellcolor{gray!20}0.6$\times$ \\
\midrule
\multirow{4}{*}{H1 (19)} & Pin. C++ & 2.9$\times$ & 1.6$\times$ & 1.6$\times$ & 2.1$\times$ & 1.2$\times$ & 1.6$\times$ \\
 & ADAM & 8.6$\times$ & 3.2$\times$ & 1.2$\times$ & 1.1$\times$ & 2.1$\times$ & 1.7$\times$ \\
 & BARD & \textbf{27}$\times$ & \textbf{22}$\times$ & 4.3$\times$ & 3.7$\times$ & 6.6$\times$ & 6.2$\times$ \\
 & BARD-c & \textbf{33}$\times$ & \textbf{26}$\times$ & \cellcolor{gray!20}1.6$\times$ & 3.7$\times$ & \cellcolor{gray!20}3.4$\times$ & \cellcolor{gray!20}1.0$\times$ \\
\midrule
\multirow{4}{*}{G1 (23)} & Pin. C++ & 2.6$\times$ & 1.6$\times$ & 1.5$\times$ & 2.0$\times$ & 1.2$\times$ & 1.6$\times$ \\
 & ADAM & --- & --- & --- & --- & --- & --- \\
 & BARD & \textbf{26}$\times$ & \textbf{21}$\times$ & 3.6$\times$ & 3.2$\times$ & 5.9$\times$ & 5.1$\times$ \\
 & BARD-c & \textbf{31}$\times$ & \textbf{23}$\times$ & 3.6$\times$ & 3.2$\times$ & 5.9$\times$ & \cellcolor{gray!20}0.8$\times$ \\
\bottomrule
\end{tabular}
\end{table}

% ================================================================
% B. SCALING PLOTS
% ================================================================
\section{Scaling Behavior Across GPUs}\label{sec:scaling_supp}

\begin{figure}[htbp]
\centering
\includegraphics[width=\textwidth]{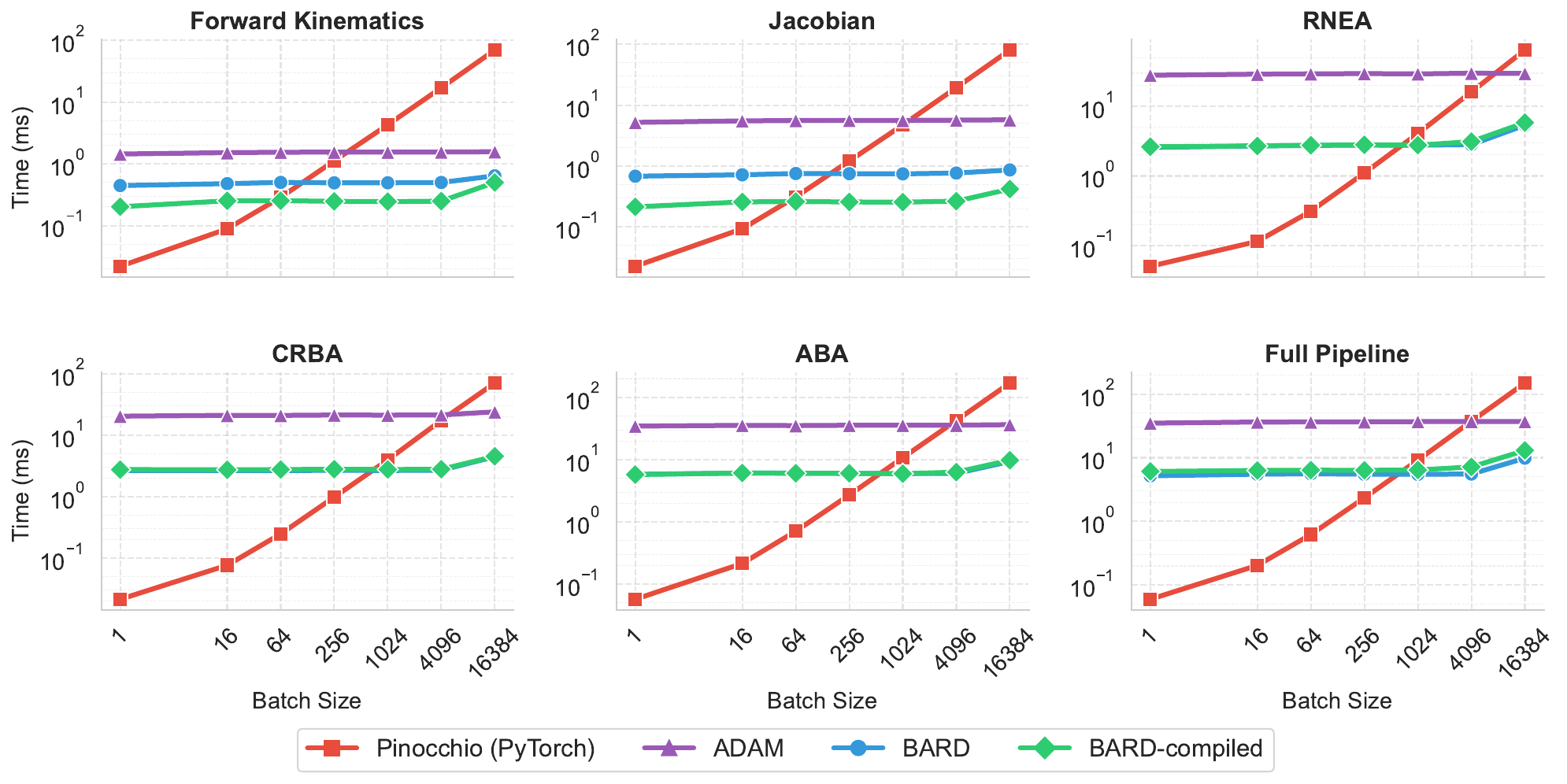}
\caption{Wall-clock time vs.\ batch size on NVIDIA H100 80GB HBM3 (Go2, 12-DOF).}\label{fig:scaling_h100}
\end{figure}

\begin{figure}[htbp]
\centering
\includegraphics[width=\textwidth]{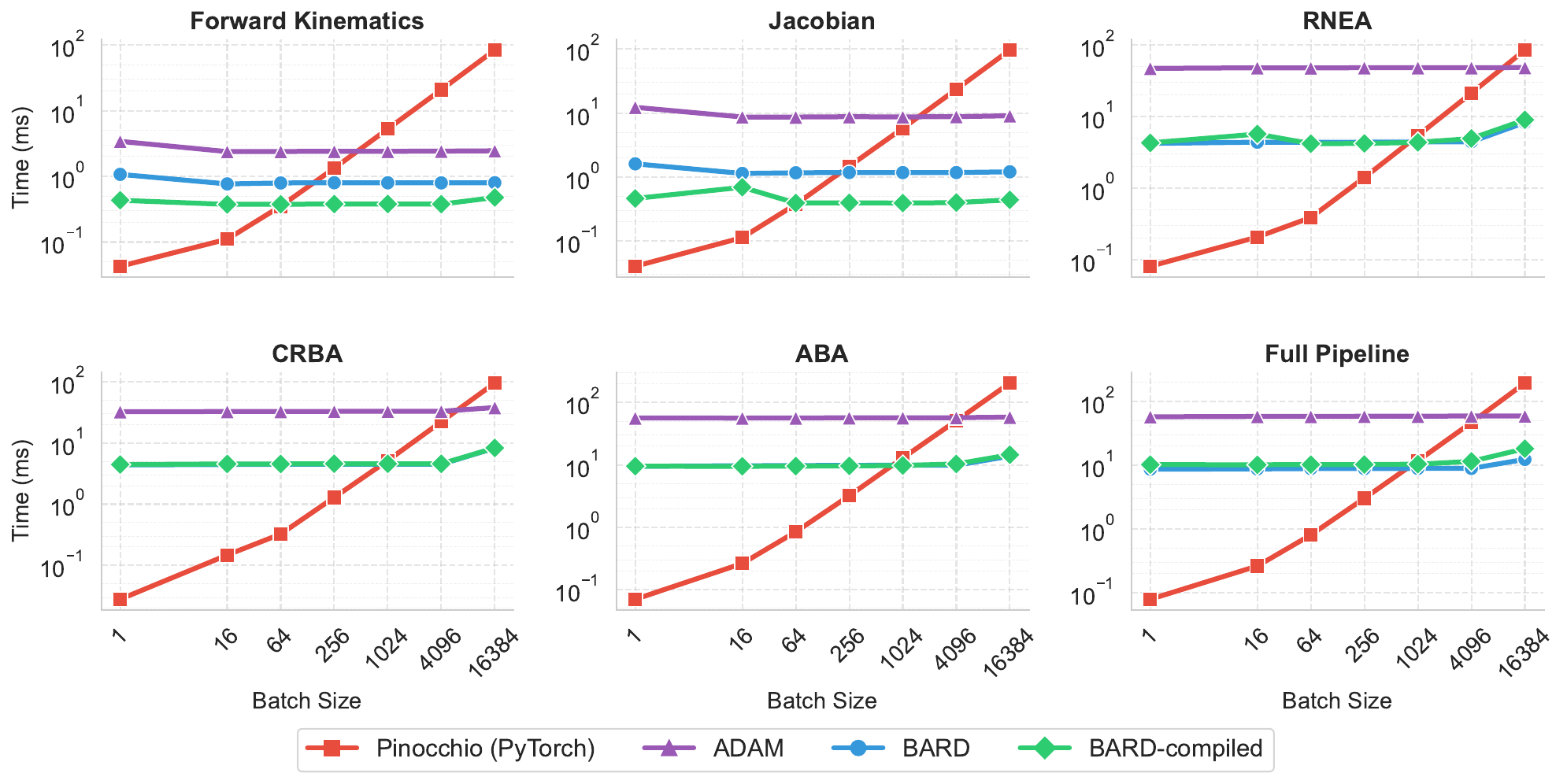}
\caption{Wall-clock time vs.\ batch size on NVIDIA A100 80GB PCIe (Go2, 12-DOF).}\label{fig:scaling_a100}
\end{figure}

\begin{figure}[htbp]
\centering
\includegraphics[width=\textwidth]{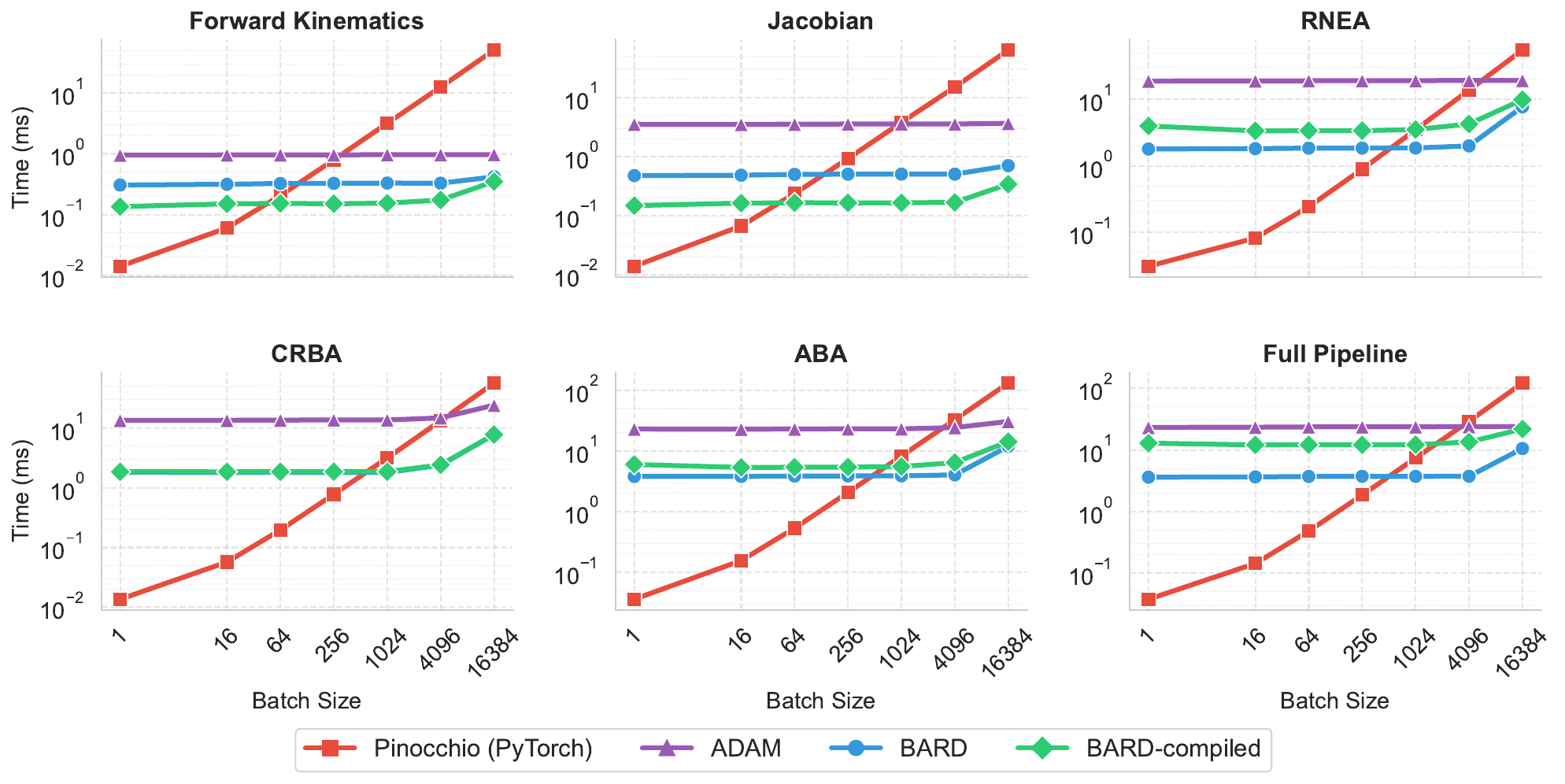}
\caption{Wall-clock time vs.\ batch size on NVIDIA L40S (Go2, 12-DOF).}\label{fig:scaling_l40s}
\end{figure}

\begin{figure}[htbp]
\centering
\includegraphics[width=\textwidth]{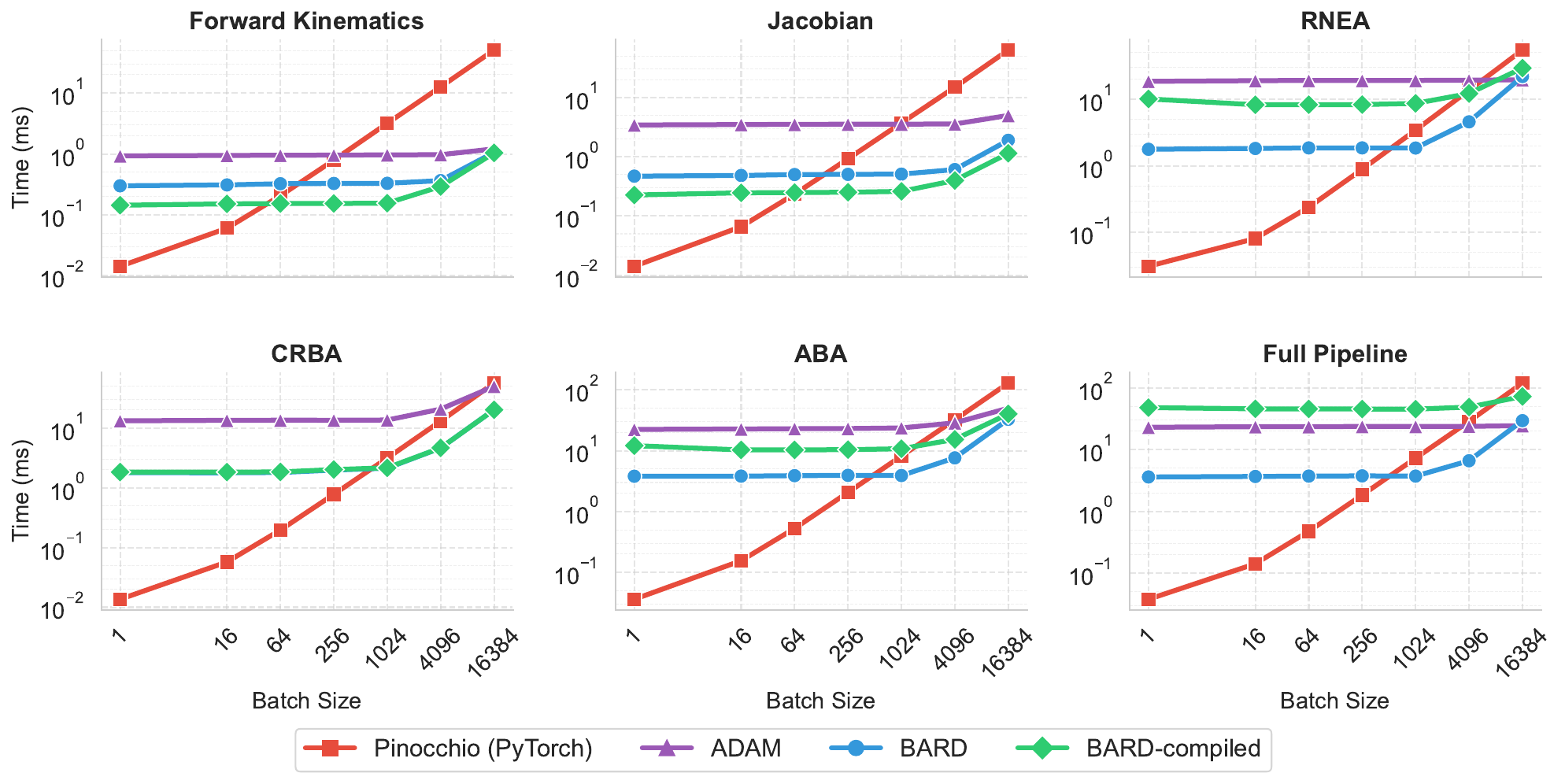}
\caption{Wall-clock time vs.\ batch size on NVIDIA L4 (Go2, 12-DOF).}\label{fig:scaling_l4}
\end{figure}

Figures~\ref{fig:scaling_h100}--\ref{fig:scaling_l4} show wall-clock time versus batch size for the Go2 robot (12-DOF) on each GPU.  On all four GPUs, \bard{}'s execution time is nearly flat from batch size 1 to 4096, after which it rises as GPU occupancy is exceeded.  Pinocchio (PyTorch) scales linearly in all cases because it processes configurations sequentially on CPU.

\FloatBarrier
% ================================================================
% C. TORCH.COMPILE ANALYSIS
% ================================================================
\section{Analysis of \texttt{torch.compile} 
Performance}\label{sec:compile}

The main text discusses the general pattern: compilation helps kinematics but can hurt dynamics on lower-end GPUs. This section quantifies the severity across all four GPUs.

On the L4, \bard{}-compiled is up to $7.4\times$ slower than eager mode for the full pipeline on the Go2 (Table~\ref{tab:speed_l4}: $0.6\times$ vs.\ $4.3\times$ over baseline). RNEA shows up to $2.7\times$ regression, and ABA up to $2.0\times$. On the L40S, the full pipeline regression reaches $3.6\times$ (Table~\ref{tab:speed_l40s}). On the H100 and A100, regression is mild and confined to the full pipeline, where compilation overhead marginally exceeds the benefit for algorithms that already have high GPU utilisation in eager mode.

The worst case is the full pipeline, which chains FK + Jacobian + RNEA + CRBA + ABA in a single compiled graph. The compiler attempts to fuse across algorithm boundaries, producing kernels too large for the L4's register file.

\end{document}